\documentclass[journal]{IEEEtran}

\ifCLASSINFOpdf
\else
   \usepackage[dvips]{graphicx}
\fi
\usepackage{url}
\usepackage{graphicx}
\usepackage{comment}
\usepackage{amsmath,amssymb} 
\usepackage{color}
\usepackage{caption}
\usepackage{diagbox}
\usepackage{stfloats}
\usepackage{float}
\usepackage{cite}

\usepackage{graphicx}
\usepackage{float}
\usepackage{color, soul}
\usepackage{authblk}
\title{SpA-Former:Transformer image shadow detection and removal via spatial attention}
\author[a]{Xiaofeng Zhang }
\author[a]{Chaochen Gu* \thanks{Corresponding author: jacygu@sjtu.edu}}
\author[a]{Shanying Zhu}

\affil[a]{School of Electronic Information and Electrical Engineering, Shanghai Jiao Tong University, Shang hai, China}
\affil[a]{Email:framebreak$\_$zxf@163.com, jacygu@sjtu.edu,shyzhu@sjtu.edu.cn}

\date{}

\usepackage{url}
\usepackage{color}
\usepackage{hyperref}
\hypersetup{
    colorlinks=true,
    linkcolor=blue,
    filecolor=blue,
    urlcolor=blue,
    citecolor=green,
}
\def\BibTeX{{\rm B\kern-.05em{\sc i\kern-.025em b}\kern-.08em
    T\kern-.1667em\lower.7ex\hbox{E}\kern-.125emX}}
\begin{document}

\maketitle

\begin{abstract}
n this paper, we propose an end-to-end SpA-Former to recover a shadow-free image from a single shaded image. Unlike traditional methods that require two steps for shadow detection and then shadow removal, the SpA-Former unifies these steps into one, which is a one-stage network capable of directly learning the mapping function between shadows and no shadows, it does not require a separate shadow detection. Thus, SpA-former is adaptable to real image de-shadowing for shadows projected on different semantic regions. SpA-Former consists of transformer layer and a series of joint Fourier transform residual blocks and two-wheel joint spatial attention. The network in this paper is able to handle the task while achieving a very fast processing efficiency. Our code is relased on \url{https://github.com/zhangbaijin/SpA-Former-shadow-removal}
\end{abstract}

\begin{figure}
\centering
\includegraphics[width=3.5in]{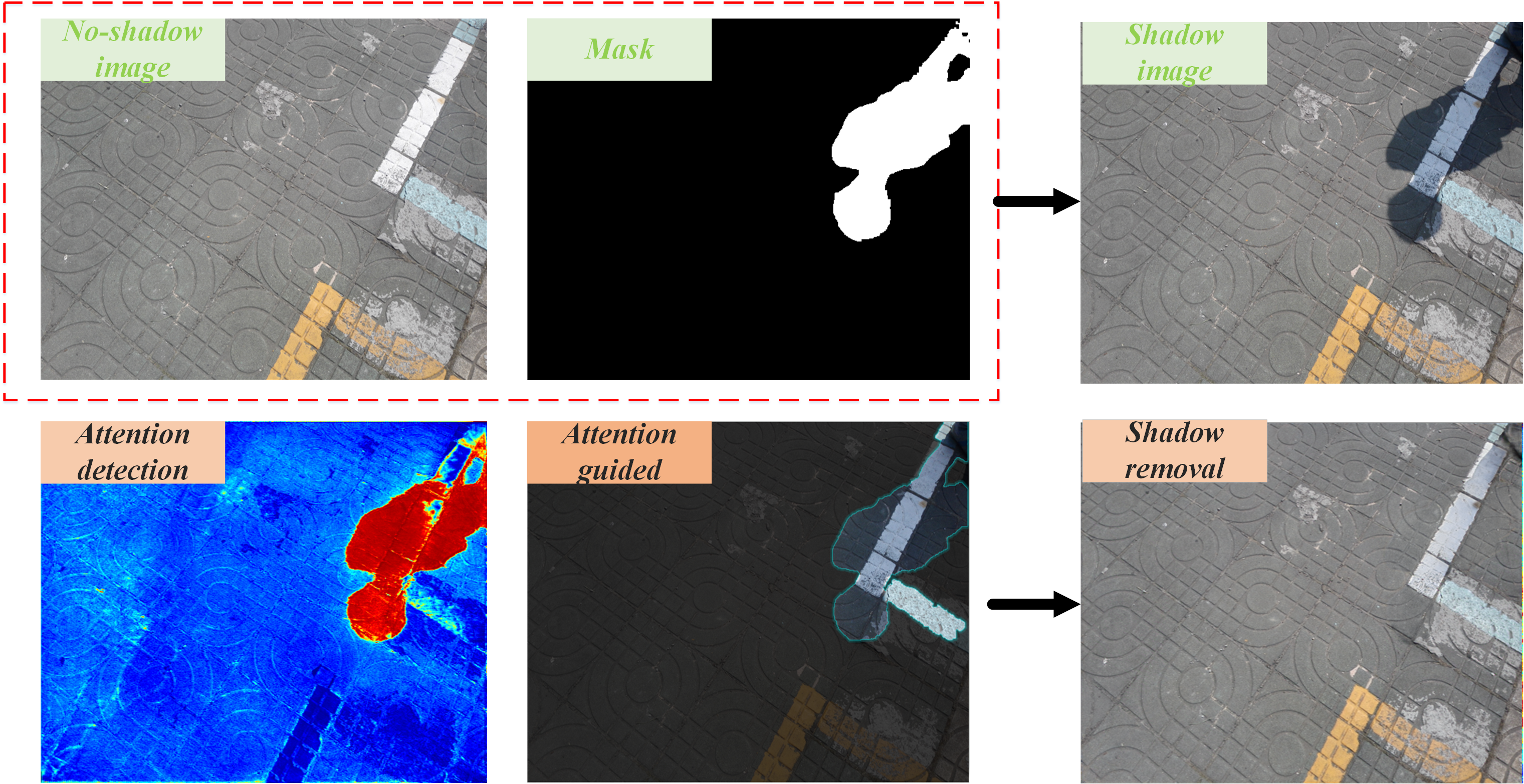}
\caption{The process of SpA-Former's shadow removal}
\label{introduction}
\end{figure}

\section{Introduction}
In natural images, the presence of shadows can provide information about the scene and lighting conditions and help us understand the scene in the image, but at the same time it makes image processing technically more difficult. In the field of image processing, shadows are usually removed, and their removal is preceded by accurate detection and localization. At present, deep learning still has problems in shadow detection, such as inconspicuous shadow boundaries and rough details; in shadow removal, the recovered shadow-free image has artifacts and changes in lighting conditions. Therefore, an effective shadow detection and removal method is urgently needed to solve the above problems. Recent studies based on deep learning have been conducted only for shadow detection or shadow removal without considering the correlation between the two tasks.
\par
It is believed that there are two main problems: First, there is much hidden knowledge about shadow removal; therefore, it is difficult to remove shadow conditions and maintain the initial image specifics, the ground occlusion information below it is completely invalid. Second, the existing network do not have the ability to consider the dependence of the remote relationship. This is because in the CNN model, the number of operations required to calculate the association increases with the distance between the two positions through convolution.
\par
 So how to remove the shadow is still a very valuable and meaningful problem. It is a common practice in end-to-end image de-shadow architectures to employ a multi-layer ResBlock, which learns the difference between a shadowed image and a clear image pair. Reconstructing a clear image from its shadow image counterpart requires changes to both low and high frequency information.
\par
To address these problems, we propose a novel Semi-supervised Transformer Network for image shadow removal named SpA-Former. It is believed that a large part of the difficulty in obtaining information about shadow occlusion is that the network adds all the functional nodes in the decoder to the global dependencies, while ignoring local information. Therefore, considering the core spatial domain information in the shadow image, we design a Two-wheel RNN joint with the spatial attention mechanism. Spatial attention can use the spatial domain information in the image as a corresponding spatial transformation so that key information can be extracted. Further more, considering that traditional ResBlock may be good at capturing the high-frequency components of an image, it tends to ignore low-frequency information. ResBlock is often unable to accurately model long-range information, which is important when reconstructing clear images from their shadow counter. Fourier transform residual module is introduced to be capable of capturing both long-term and short-term interactions while integrating low and high frequency residual information.
\par
Last but not least, in the existing deep learning shadow removal algorithms, seldom consider the relationship of remote dependencies in the network, and learning remote dependencies is a key challenge. A key factor affecting the ability to learn this dependence is the path length that the forward and backward signals must traverse in the network. The shorter these paths between any combination of positions in the input and output sequences, the easier it is to learn distance dependencies. For this reason, we designed the Transformer network, which contains contexture extractors and correlation embedding modules to increase the dependency of the network.

In summary, our contributions are the following three:
\begin{enumerate}
 \item In order to remove shadows effectively, this paper designs a one-stage ensemble shadow detection and shadow removal network named SpA-Former, while this paper is the first to introduce transformer as an image de-shadowing method.
 \item This paper proposes a joint Fourier transform residual block and two-wheel joint spatial attention group, converting the spatial domain information into an image, while considering the relationship of remote dependencies.
\item The network designed in this paper is a one-stage network that combines shadow detection and shadow removal, and achieves satisfactory results in both shadow detection and shadow removal.
\end{enumerate}
\section{Realted work}

\subsection{Supervised learning image shadow removal}
The earliest weakly supervised first for DeshadowNet \cite{1} has the biggest feature of fully automatic end-to-end implementation of shadow removal, while the biggest contribution of this paper is to propose a dataset SRD (A New Dataset for Shadow Removal), automatic and end-to-end deep neural network (DeshadowNet), the proposed DeshadowNet is multi-contextual, which combines high-level semantic information, mid-level appearance information and local image details to perform the final prediction. The authors of the second ST-CGAN \cite{2} propose a multitasking perspective that differs from all existing approaches in that it learns detection and elimination jointly in an end-to-end manner with the aim of jointly exploiting the advantages of each other's improvements. The framework is based on a new "Stacked Conditional Generative Adversarial Network (ST-CGAN)", which consists of two stacked Conditional Generative Adversarial Networks (CGANs), each with a generator and a discriminator, while another big contribution of ST-CGAN is the proposed dataset ISTD. \par
SID \cite{3} designs depth networks to illuminate the shaded regions by estimating a linear transformation function. For penumbra (half-shadow), the shadow matting technique is used to handle it. The shadow regions are relit (lit), replacing the shaded regions in the shadow image. The fourth DSC \cite{4}develops direction-aware attention mechanisms in spatial recurrent neural networks (RNNs) by introducing attention weights when aggregating spatial contextual features in the RNN. By training to learn these weights, we can recover the direction-aware spatial context (DSC) to detect and eliminate shadows. The design is developed into a DSC module and embedded into a convolutional neural network (CNN) to learn different levels of DSC functions. RSI-GAN \cite{5}designs a general framework to mine the information of residual and illumination through multiple GANs for shadow removal. DHAN \cite{6} uses a dual-level aggregation network (DHAN) in order to eliminate boundary artifacts. DHAN does not have any downsampling and consists of a series of growing inflated convolutions as a backbone for attention and prediction by aggregating multiple layers of contextual features. Auto-exposure \cite{7} uses a fusion exposure approach, where the exposure parameters of the shadow part are predicted by a network with regular dithering of the exposure, and a fusion network is used to Learning the fusion parameters, the previous results are automatically fused to obtain a shadow-free image, and the boundary residues are further removed by a RefineNet.
\par
CANet \cite{8} aims to mine the contextual information of the shadowed and non-shadowed regions. It is not just removal without detection, he relies on the global luminance averaging method to achieve the detection, the image from RGB to LAB only L channel is sensitive to shadows, the L here for global averaging, you can get a kind of fuzzy shadow-free map, you can distinguish between shadow-free and shadowed areas. Because the previous approaches of deep learning focus on increasing the perceptual field of the model and ignore the information of pairwise matching in the image.
\subsection{Weakly supervised image shadow removal}
Mask-ShadowGAN \cite{9} argues that previous solutions to shadow removal problems using deep learning are supervised, paired Data. using paired, there is no drawback for shadow removal methods. But the process of obtaining the paired dataset can be problematic.
Therefore Mask-shadowGAN uses the idea of using CycleGAN, using unpaired data to train the model, guiding the generation of shadow images by shadow mask, solving the problem of multiple shadow maps corresponding to a shadow-free map, and modeling the relationship between shadow and shadow-free by the difference between shadow and shadow-free.LG- ShadowNet \cite{10} argued that in practice, CNN training considers the ease of training and unpaired data is more favored for data collection. Therefore, they proposed a new Lightness-Guided Shadow Removal Network (LG-ShadowNet \cite{10}) capable of training on unpaired data.
\subsection{Unsupervised learning of image shadow removal}
G2R \cite{11} they exploited the fact that shaded images usually contain both shaded and non-shaded regions. By this method, a set of shaded and unshaded patches can be cropped to construct unpaired data for network training,proposing three sub-network modules: shadow generation, shadow removal, and shadow refinement. TC-GAN \cite{12} performs the shadow removal task in an unsupervised manner. Comparing the GAN-based unsupervised shadow removal method with the bidirectional mapping in cyclic consistency, TC-GAN aims to learn a unidirectional mapping that converts the late-shadowed image into a shadow-free image. By the proposed goal consistency constraint aimed at connecting two GAN-based sub-networks, the correlation is between the shaded image and the output unshaded image, and the authenticity of the recovered unshaded image is strictly
constrained.
\begin{figure*}
\centering
\includegraphics[width=6in]{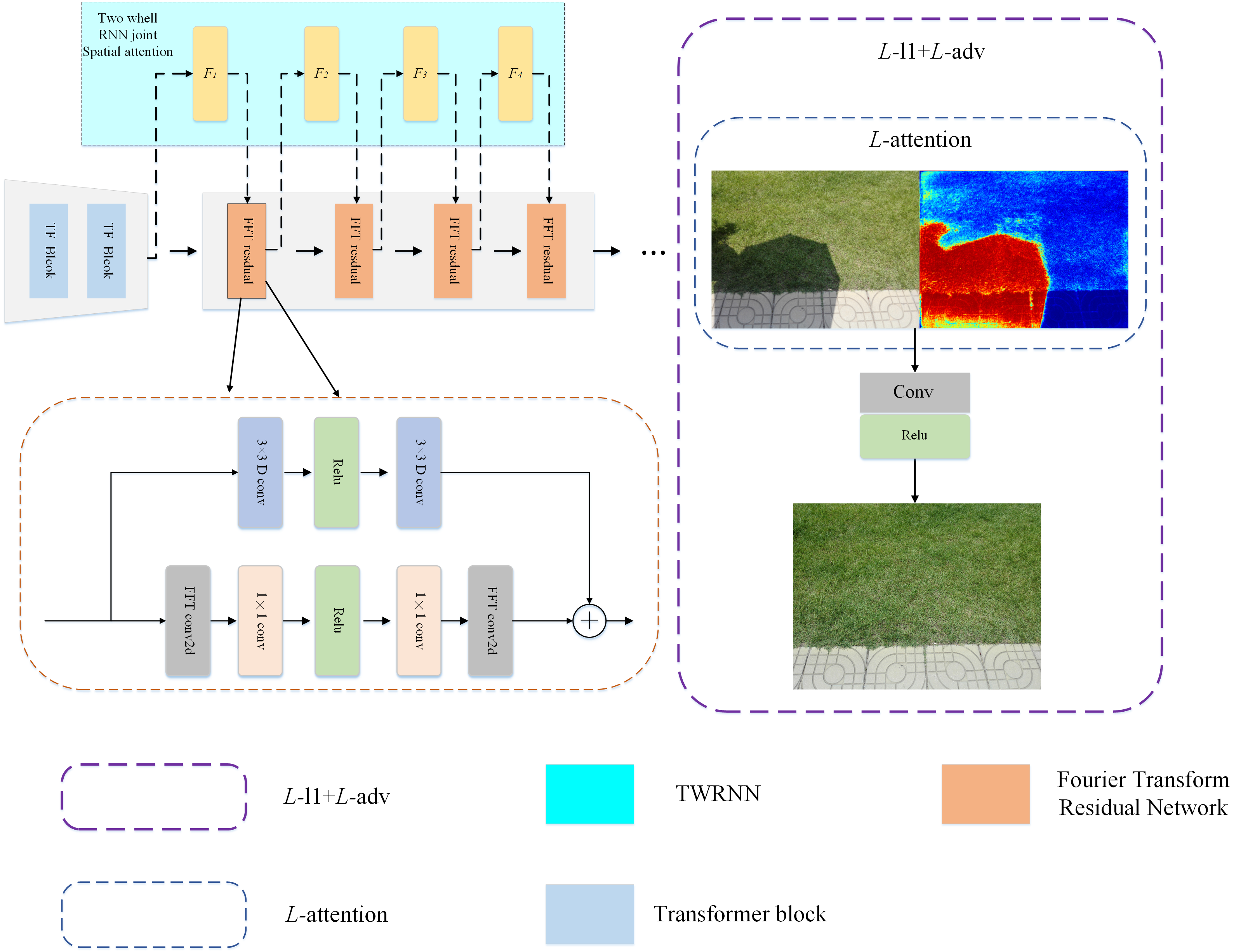}
\caption{The structure of SpA-Foremr }
\label{structure}
\end{figure*}

\section{Methodology}
 The structure of SpA-Former is shown in Fig \ref{structure}.  SpA-Former is enspired by \cite{18,19}, it consists of a deep Transformer and CNN network. First of all, from the Transformer network, then the feature map is extracted by 3$\times$3 convolution, following bottlenet and Two-Wheel RNN Joint Spatial Attention, TWRNN(Two-Wheel RNN Joint Spatial Attention) module is designed to make the network pay attention to specific cloud images, it can discover and find the focus map from the input element map. The attention graph is a two-dimensional matrix, in which the value of each element is a continuous value, indicating how much attention should be assigned to the pixel.

\subsection{Transformer layer }
Transformer is able to capture the global and local relationships in one step, and does not have the limitation of sequence length for capturing long-term dependencies as RNNs do. It also complements Two-Wheel RNN Joint Spatial Attention network in the next section, where the result of each step does not depend on the previous step and can be made into a parallel pattern. This enables the network to perform joint feature learning on real images and shadow images, and can capture more accurate texture features.
\par
Starting from the high-resolution input, the encoder hierarchically reduces spatial size, while expanding channel capacity, we let Transformer blocks to aggregate the
low-level image features of the encoder with the high-level features of the decoder. It is beneficial in preserving the fine structural and textural details in the restored images. Next, the deep features are further enriched in the re-finement stage operating at high spatial resolution.
\par
The main computational overhead of Transformers comes from the self-attentive layer. The time and memory complexity of querying dot product interactions in traditional self attention keys grows quadratically with the spatial resolution of the input, i.e., for images of $W\times H$ pixels, and therefore it is not feasible to apply SA to most image recovery tasks that typically involve high resolution images.
To alleviate this problem, we propose Transformer blocks with linear complexity. the key component is to apply self attention across channels rather than spatial dimensions, i.e., to compute the cross-covariance across channels generate an attention graph encoding the global context implicitly. Deep convolution is introduced to emphasize the local context global attention map before computing feature covariance to generate from a layer normalized tensor $\mathbf{Y} \in \mathbb{R}^{\hat{H} \times \hat{W} \times \hat{C}}$, our Transformer block first generates query (Q), key (K), and value (V) predictions that enrich the local context. This is achieved by applying $1\times1$ convolution to aggregate pixel-by-pixel cross-channel context, followed by $3\times3$ deep convolution to encode channel space context, producing  $\mathbf{Q}=W_{d}^{Q} W_{p}^{Q} \mathbf{Y}, \mathbf{K}=W_{d}^{K} W_{p}^{K} \mathbf{Y} \text { and } \mathbf{V}=W_{d}^{V} W_{p}^{V} \mathbf{Y}$. Where $Wp()$ is a $1×1$ point-by-point convolution and $Wd()$ is a $3\times3$ deep convolution. We use the layers in the unbiased convolutional network. Next, we reshape the query and key projections so that their dot product interactions generate transposed attention of size $R^(C)(^C)$. Overall, the process is defined as follows:

$\hat{\mathbf{X}}=W_{p}$ Attention $(\hat{\mathbf{Q}}, \hat{\mathbf{K}}, \hat{\mathbf{V}})+\mathbf{X}$
$\operatorname{Attention}(\hat{\mathbf{Q}}, \hat{\mathbf{K}}, \hat{\mathbf{V}})=\hat{\mathbf{V}} \cdot \operatorname{Softmax}(\hat{\mathbf{K}} \cdot \hat{\mathbf{Q}} / \alpha)$

Here, $\alpha$ is a learnable scaling parameter to control the magnitude of the dot product of K and Q before applying the softmax function.
\subsection{Two-Wheel RNN joint spatial Attention}
\begin{figure}
\centering
\includegraphics[width=3.5in]{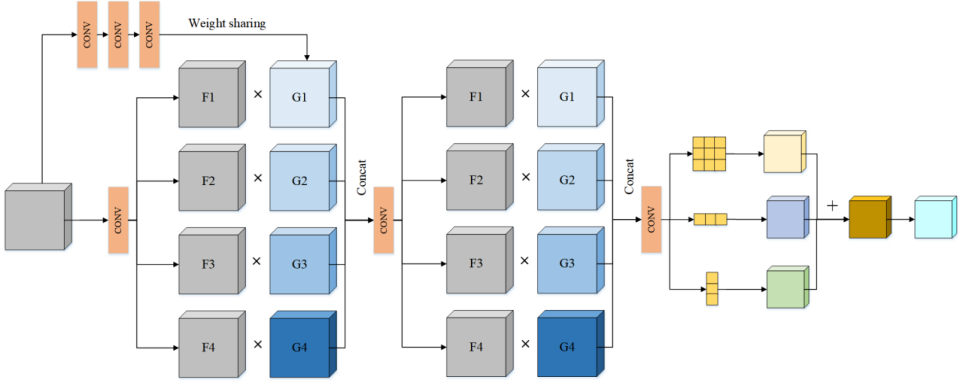}
\caption{The structure of TRNN}
\label{TRNN}
\end{figure}

Two-Wheel RNN Joint Spatial Attention model is builded based on the above two-round four-way RNN architecture. RNN model is used to project the descent in four main directions. As shown in Fig. \ref{TRNN}, another branch has been added to capture spatial context information to selectively highlight the expected shadow features. TWRNN can effectively identify areas affected by clouds according to the input shadow image. Three standard residual blocks are first used to extract features to guide the three subsequent attention residual blocks to eliminate shadows by learning negative residual. Finally, the generated feature map is fed into two standard residual blocks to reconstruct the final shadow-removed image. In the next steps, the iterative update gradually focuses the attention on all shaded areas, and the shaded areas marked in red are marked more and more accurately in the next steps.

$F_1,F_2,F_3,F_4$ represent the eigenvalues in the four directions (up, down, left and right), while $G_1,G_2,G_3,G_4$, represents the weight matrix corresponding to the four directions (up, down, left and right).

\subsection{Joint Fourier Transform Residuals Module}
FTR is introduced by \cite{20}, it is a common practice in end-to-end image recovery architectures to employ ResBlock, which learns the difference between blurred and clear image pairs. Reconstruction of a non-shadowed image from a shaded counterpart requires changes to both low and high frequency information. While traditional ResBlock may be good at capturing the high-frequency components of an image, it tends to ignore low-frequency information. In addition, ResBlock is often unable to accurately model long-range information, which is important when reconstructing unshaded images from their shadowed counterparts. The residual information is included to obtain the detected attention map by multiplication, and then it is used to obtain the negative residual, which is used to restore the image of the previous step to a slightly shaded or unshaded image. As the result graph shows, the shaded part of the output image becomes lighter and lighter as the progressive steps are taken, and the final output is almost free of shaded parts.

\subsection{Loss function}

The loss function is mainly composed of five items: ${L_{CGAN}}$, ${L_{1}}$, and ${L_{attention}}$.
\begin{equation}
\begin{split}
L_{CG A N}(G, D)=\mathcal{E}_{x, y \sim p_{d d a}(x, y)}[\log D(x, y)]+ \\
\mathcal{E}_{x \sim p_{d a t a}(x), z \sim p_{z}(z)}[\log (1-D(x, G(x, z)))]
\end{split}
\end{equation}

The second part of the loss is the standard $L_{1}$ loss, which is used to measure the accuracy of each reconstructed pixel. As shown in (5), $I_{input}$ and $I_{output}$ are the input and output images, respectively, $\lambda_{C}$ is the weight of each channel, $\phi$ is the predicted result of the network, and $C$, $H$ and $W$ indicate the number of channels, the height and the width of the image, respectively.
\begin{equation}
\begin{split}
L_{1}(G)=\frac{1}{4 H W} \sum_{c=1}^{C} \sum_{v=1}^{H} \sum_{u=1}^{W} \lambda_{c}\left|I_{\text {outpud}}^{(u, v, c)}
-\phi\left(I_{\text {input}}\right)^{(u, v, c)}\right|_{1}
\end{split}
\end{equation}
The third part of the loss is attention loss, which is defined as (6). The matrix $A$ is the attention map module generated by soft attention, and matrix $M$ is the binary image of the cloud area, which is composed of cloudy and cloudless images.
\begin{equation}
L_{Attention}=\|A-M\|_{2}^{2}
\end{equation}

The total loss function is:
\begin{equation}
\begin{split}
L_{total}=& L_{1}+ L_{CGAN}+ L_{Attention}
\end{split}
\end{equation}

\section{Experiment}
\subsection{Dataset and Experimental setup }
\begin{figure}[h]
\centering
\includegraphics[width=3in]{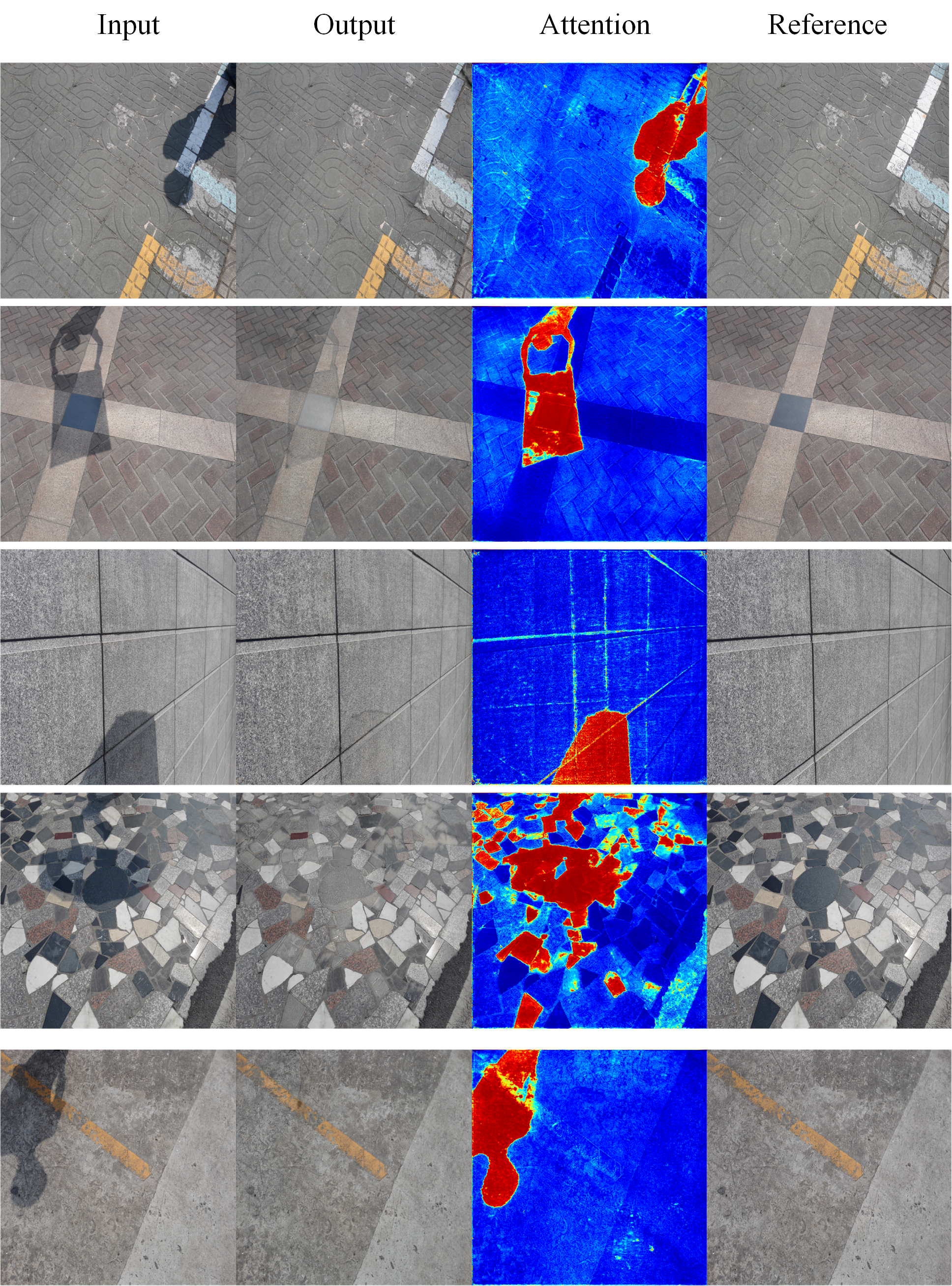}
\caption{Results of SpT-former}
\label{result}
\end{figure}

This paper uses the latter dataset ISTD \cite{2} (Large-scale Dataset with Image Shadow Triplets), which differs from other datasets containing shadow maps in that the ISTD dataset contains three types of data, the original image containing shadows, the shadow labeled data and the original image not containing shadows, and contains 1870 triples, suitable for multi-task training. It is randomly divided into 1330 training and 540 testing. the learning rate was set to 0.0004, the small batch was 1, and the period was 200. We chose the Adam optimizer with a $brta$ of 0.999 at the same time. The results of SpA-Former is shown in \ref{result}.

\subsection{ Performance Comparisons}
Our method is compared with existing methods including Yang \cite{17}, Guo \cite{15}, Gong \cite{16}, DeShadowNet\cite{1}, STC-GAN \cite{2}, DSC \cite{4}, Mask-ShadowGAN\cite{9}, RIS-GAN \cite{5}, DHAN \cite{6}, SID \cite{3}, LG-shadow \cite{10}, G2R\cite{11}. We adopt the root mean square error (RMSE), structure similarity index (SSIM) and Peak Signal to Noise Ratio(PSNR) in the LAB color space as evaluation metrics. While RMSE directly measures the per-pixel error between the recovered images and the ground truth image. Table 1 report the RMSE, SSIM and PSNR values, respectively, of different shadow removal methods on the ISTD dataset \cite{2}. We evaluate the performance of different methods on the shadow regions, non-shadow regions, and the whole image. We can see that the proposed SpA-former achieves the satisfactory performance among all the compared methods.
\par
By the way, the current image de-shadowing networks are roughly divided into three categories, weakly supervised, supervised, and unsupervised. Regardless of which one, the current common method is the two-stage method, which first detects and then removes the shadow part, however, the training time and GPU performance required for the two-stage method are relatively high. Our proposed SpA-former belongs to the one-stage network, and he is able to remove the shadow part while recognizing it through the spatial attention map, and more importantly, the network in this paper can be trained and used on an ordinary 1080Ti, which greatly reduces the application barriers.

\begin{table*}[t]\caption{Performance comparison of shadow removal on ISTD}
\centering
\resizebox{15cm}{2.5cm}{
\begin{tabular}{c|c|c|c|c|c|c|c|c|c}
\hline
Models &  RMSE  & RMSE-N& RMSE-S& SSIM  &SSIM-N& SSIM-S &PSNR  & PSNR-N & PSNR-S\\
\hline Yang[TIP2012\cite{17}]  & \ 15.63   &  14.83  & 19.82 & -  & - & - & - & - & -    \\
\hline Guo[TPAMI2013\cite{15}]  &   9.3 & 7.46   & 18.95 &0.919 &0.944 &0.978 &23.07 &24.86 &30.98 \\
\hline Gong[BMVC2014\cite{16}] & 8.53 & 7.29 & 14.98 &0.908 &0.929 &0.98 &24.07 &25.26 &32.43\\
\hline DeShadowNet[CVPR2017\cite{1}]  & 7.83 & 7.19 & 12.76 & -  & - & - & - & - & -  \\
\hline STC-GAN[CVPR2018\cite{2}]  & 7.47 & 6.93 &10.33 &0.929 &0.947 &0.985 &27.43 &28.67 &\textbf{35.8}\\
\hline DSC[TPAMI2018\cite{4}] & 6.67 & \ 6.39& 9.22  &0.845 &0.885 &0.967 &26.62 &28.18 &33.45\\
\hline Mask-ShadowGAN[ICCV2019\cite{9}]  &7.61	&7.03	&10.35 &-  & - & - & - & - & -  \\
\hline RIS-GAN[AAAI2019\cite{5}]  &6.62	&6.31	&9.15 &-  & - & - & - & - & - \\
\hline DHAN[CVPR2019\cite{6}]  &\textbf{6.28}	&\textbf{5.92}	&\textbf{8.43}	&0.921	&0.941	&0.983	&27.71	&29.54	&34.79\\
\hline  SID[ICCV2019\cite{3}]  &7.96	&7.72	&9.64	&\textbf{0.948}	&0.964	&\textbf{0.986}	&25.01	&26.1	&32.88\\
\hline LG-shadow[ECCV2020\cite{10}]&6.67	&5.93	&11.51	&0.906	&0.938	&0.974	&25.83	&28.32	&31.08\\
\hline G2R[CVPR2021\cite{11}]  &7.84	&7.54  &10.71 &0.932 &\textbf{0.967}	&0.974	&24.72	&26.18	&31.62\\
\hline Our &6.86	& 6.22	&10.48	&0.931	& 0.956	&0.982	&\textbf{27.73}	&\textbf{30.16} &33.51\\
\hline
\end{tabular}}
\label{table1}
\end{table*}

\begin{figure*}[t]
\centering
\includegraphics[width=7in]{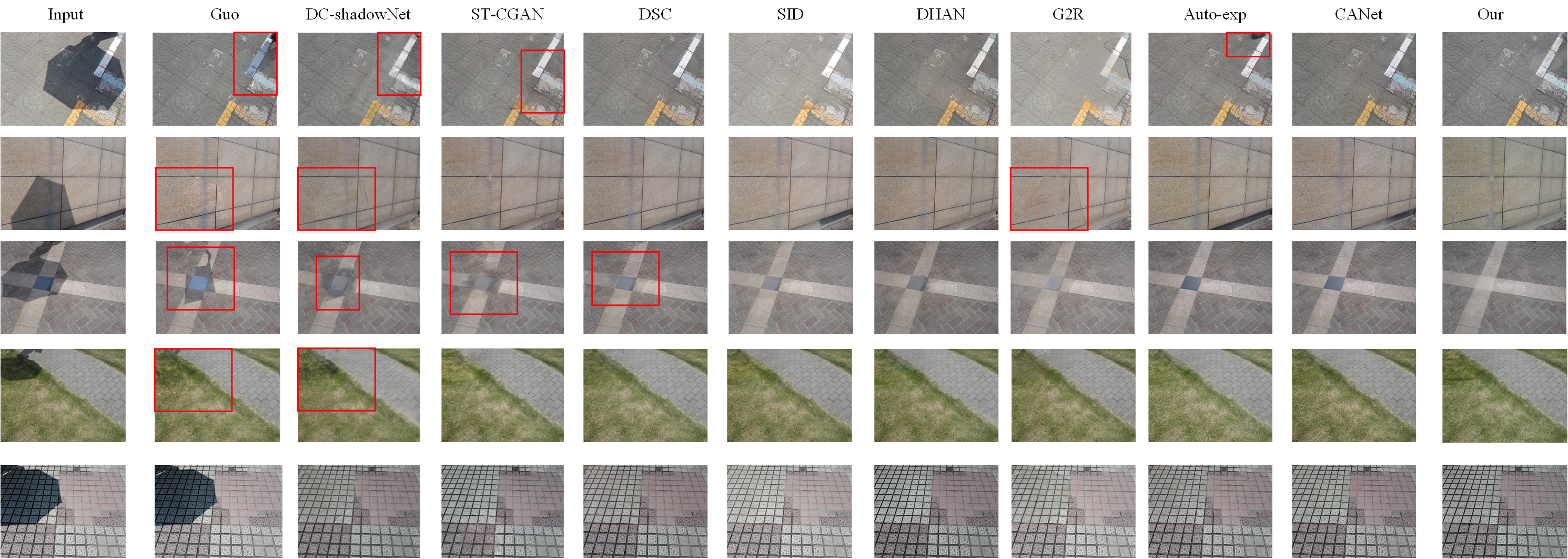}
\caption{The compare results of SpA-Former with other exsiting methods}
\label{compare}
\end{figure*}

\begin{figure}
\centering
\includegraphics[width=3.5in]{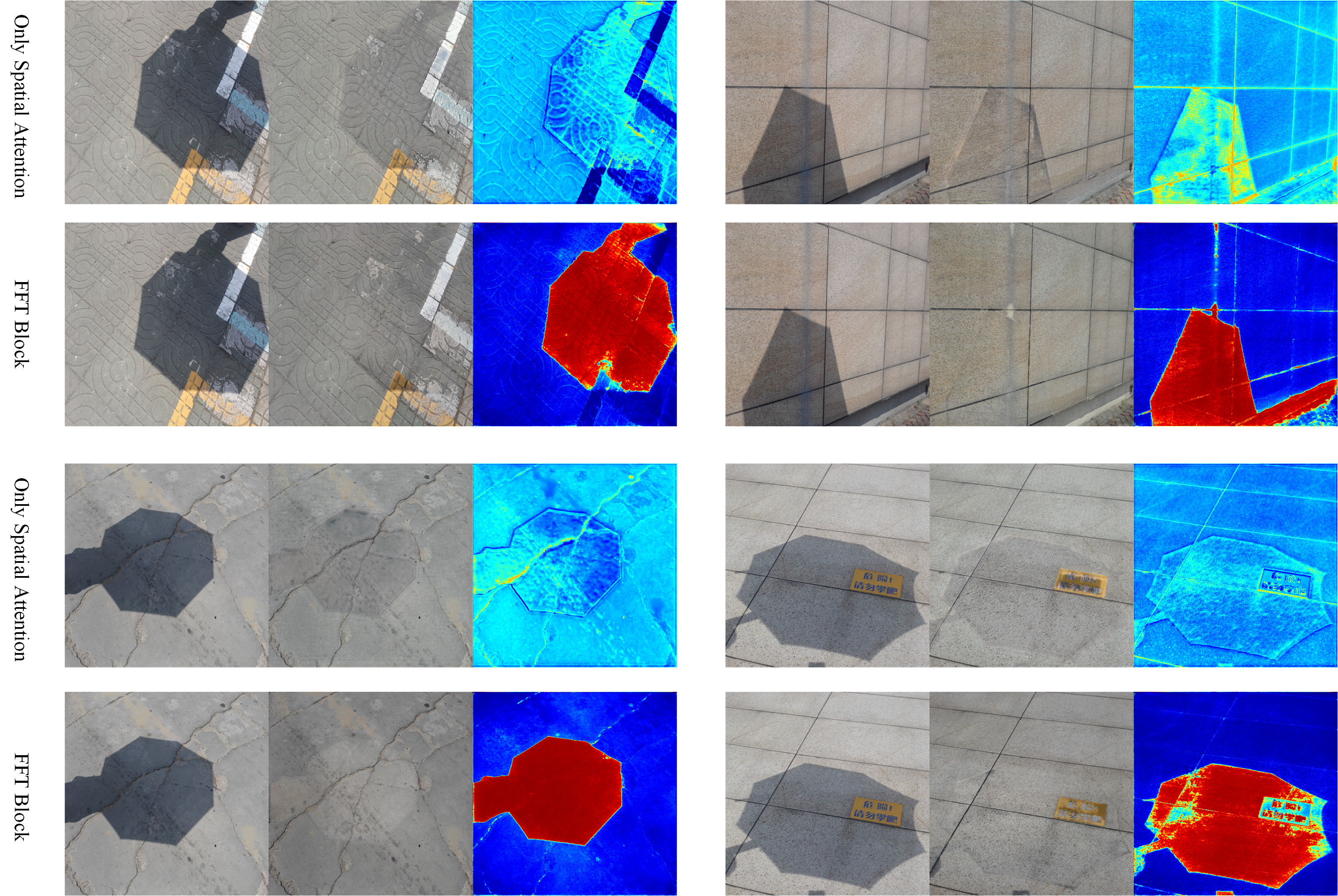}
\caption{Ablation of FTR Block}
\label{ablation1}
\end{figure}

\begin{figure}
\centering
\includegraphics[width=3.5in]{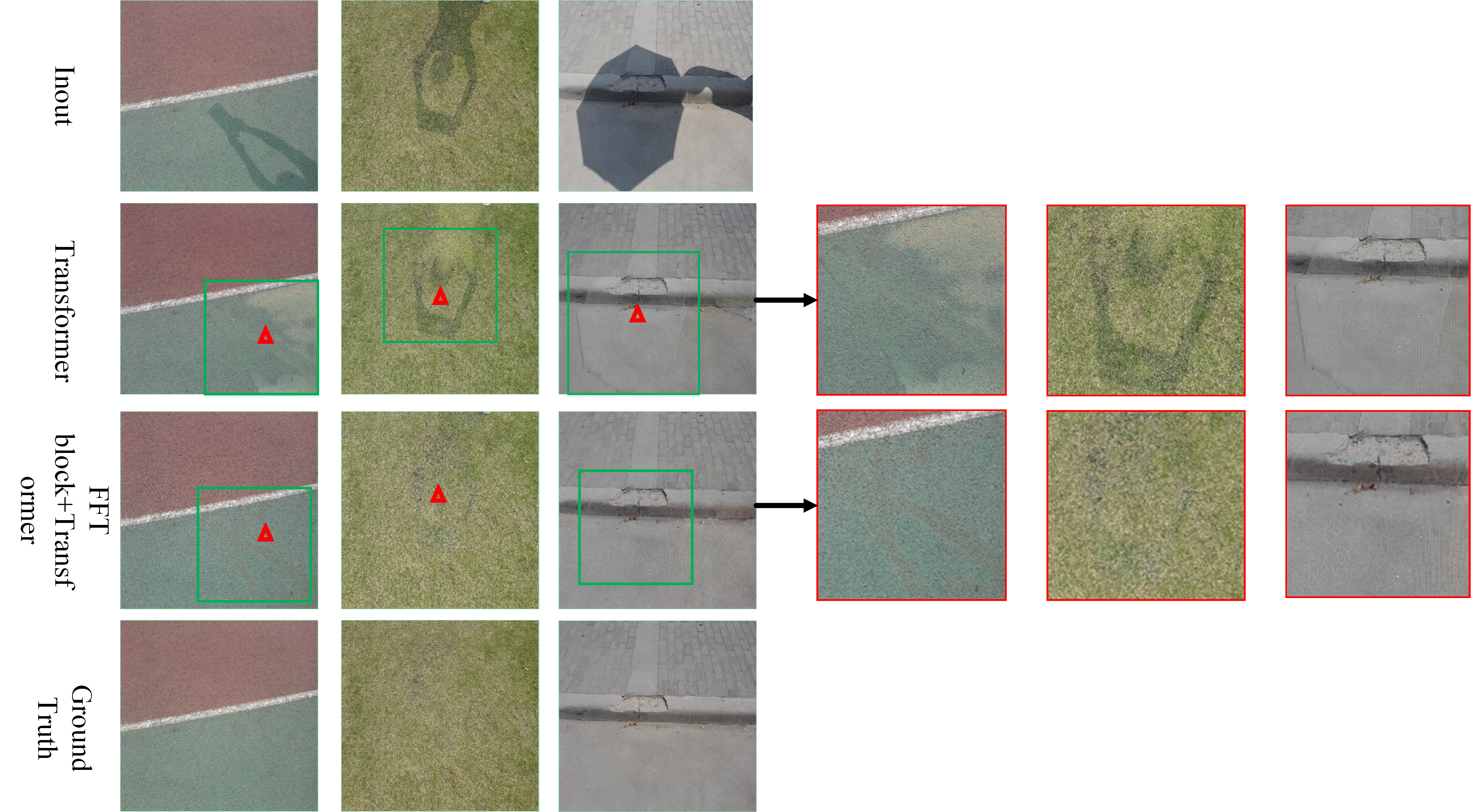}
\caption{Ablation of FTR and Transformer}
\label{ablation3}
\end{figure}

\subsection{Ablation Study on SpA-Former}

To validate the effectiveness of Transformer, Two-wheel RNN joint spatial attention and FTR Block in our network, we make the ablation experiment in Table II. Only using Transformer or Two-wheel RNN could get a significant improvement over classical encoder-decoder. And the combination of them could achieve the best performance, indicating the
regulation is beneficial for the task. We also visualize the results of ISTD to measure the quantitative capability of them. As shown in Fig. 3, the combination of these proposed two parts could recover the details and generate realistic results in easy nature and complicated ISTD cases.

\begin{table*}[t]\caption{Performance comparison of shadow removal on ISTD}
\centering
\resizebox{15cm}{1cm}{
\begin{tabular}{c|c|c|c|c|c|c|c|c|c}
\hline
 &  RMSE  & RMSE-N& RMSE-S& SSIM  &SSIM-N& SSIM-S &PSNR  & PSNR-N & PSNR-S\\
\hline  Baseline &  7.87   &  7.19  & 12 & 0.911  & 0.939 & 0.970 & 26.21 & 29.10 & 31.65    \\
\hline  Added FRB &   7.04 & 6.34   & 11.07 &0.92 &0.949 &0.979 &26.87 &29.59 &32.34 \\
\hline Added Tansformer & 7.03 &6.18 &9.50  &0.884 &0.915 &0.974 &26.82 &29.67 &31.78\\
\hline  Added Transformer +FRB & 6.86 & 6.22 & 10.48 &0.931 &0.956 &0.982 &27.73 &30.16 &33.51\\
\hline
\end{tabular}}
\label{comparetable}
\end{table*}

\section{Results and discussion}
In this paper, we propose an end-to-end SpA-Former to recover a shadow-free image from a single shaded image. Unlike traditional methods that require two steps for shadow detection and then shadow removal, the SpA-Former unifies these steps into one, which is a one-stage network capable of directly learning the mapping function between shadows and no shadows, it does not require a separate shadow detection step, nor does it have any post-processing refinement step. Thus, SpA-former is adaptable to real image de-shadowing for shadows projected on different semantic regions. SpA-Former consists of transformer layer and a series of joint Fourier transform residual blocks and two-wheel joint spatial attention. The network in this paper is able to handle the task while achieving a very fast processing efficiency. In the future, we will adapt and extend this variation task (such as rain and snow removal).


\begin{thebibliography}{99}

\bibitem{1}Qu L, Tian J, He S, et al. Deshadownet: A multi-context embedding deep network for shadow removal[C]//Proceedings of the IEEE Conference on Computer Vision and Pattern Recognition. 2017: 4067-4075.
\bibitem{2}Wang J, Li X, Yang J. Stacked conditional generative adversarial networks for jointly learning shadow detection and shadow removal[C]//Proceedings of the IEEE Conference on Computer Vision and Pattern Recognition. 2018: 1788-1797.
\bibitem{3}Le H, Samaras D. Shadow removal via shadow image decomposition[C]//Proceedings of the IEEE/CVF International Conference on Computer Vision. 2019: 8578-8587.
\bibitem{4}Hu X, Fu C W, Zhu L, et al. Direction-aware spatial context features for shadow detection and removal[J]. IEEE transactions on pattern analysis and machine intelligence, 2019, 42(11): 2795-2808.
\bibitem{5}Zhang L, Long C, Zhang X, et al. Ris-gan: Explore residual and illumination with generative adversarial networks for shadow removal[C]//Proceedings of the AAAI Conference on Artificial Intelligence. 2020, 34(07): 12829-12836.
\bibitem{6}Cun X, Pun C M, Shi C. Towards ghost-free shadow removal via dual hierarchical aggregation network and shadow matting GAN[C]//Proceedings of the AAAI Conference on Artificial Intelligence. 2020, 34(07): 10680-10687.
\bibitem{7}Fu L, Zhou C, Guo Q, et al. Auto-exposure fusion for single-image shadow removal[C]//Proceedings of the IEEE/CVF Conference on Computer Vision and Pattern Recognition. 2021: 10571-10580.
\bibitem{8}Chen Z, Long C, Zhang L, et al. CANet: A Context-Aware Network for Shadow Removal[C]//Proceedings of the IEEE/CVF International Conference on Computer Vision. 2021: 4743-4752.
\bibitem{9}Hu X, Jiang Y, Fu C W, et al. Mask-ShadowGAN: Learning to remove shadows from unpaired data[C]//Proceedings of the IEEE/CVF International Conference on Computer Vision. 2019: 2472-2481.
\bibitem{10}Liu Z, Yin H, Mi Y, et al. Shadow removal by a lightness-guided network with training on unpaired data[J]. IEEE Transactions on Image Processing, 2021, 30: 1853-1865.
\bibitem{11}Vasluianu F A, Romero A, Van Gool L, et al. Self-Supervised Shadow Removal[J]. arXiv preprint arXiv:2010.11619, 2020.
\bibitem{12}Le H, Samaras D. From shadow segmentation to shadow removal[C]//European Conference on Computer Vision. Springer, Cham, 2020: 264-281.
\bibitem{13}Liu Z, Yin H, Wu X, et al. From shadow generation to shadow removal[C]//Proceedings of the IEEE/CVF Conference on Computer Vision and Pattern Recognition. 2021: 4927-4936.
\bibitem{14}Tan C, Feng X. Unsupervised Shadow Removal Using Target Consistency Generative Adversarial Network[J]. arXiv preprint arXiv:2010.01291, 2020.
\bibitem{15}R. Guo, Q. Dai, and D. Hoiem, ¡°Paired regions for shadow detection and removal,¡± IEEE TPAMI, vol. 35, no. 12, 2012.
\bibitem{16}H. Gong and D. Cosker, ¡°Interactive shadow removal and ground truth for variable scene categories,¡± in Proc. BMVC, 2014.
\bibitem{17}Q. Yang, K. Tan, and N. Ahuja. Shadow removal using bilateral filtering. IEEE TIP, 21(10):4361¨C4368, 2012.
\bibitem{18}Pan H. Cloud removal for remote sensing imagery via spatial attention generative adversarial network[J]. arXiv preprint arXiv:2009.13015, 2020.
\bibitem{19}Zamir S W, Arora A, Khan S, et al. Restormer: Efficient transformer for high-resolution image restoration[C]//Proceedings of the IEEE/CVF Conference on Computer Vision and Pattern Recognition. 2022: 5728-5739.
\bibitem{20}Mao X, Liu Y, Shen W, et al. Deep residual fourier transformation for single image deblurring[J]. arXiv preprint arXiv:2111.11745, 2021.

\end{thebibliography}
\end{document}